\documentclass{article}
\usepackage{spconf,amsmath,graphicx}
\usepackage{amsmath,graphicx, url}
\usepackage{makecell}
\usepackage{multirow}
\usepackage{arydshln}

\DeclareMathOperator*{\argmax}{arg\,max}

\title{MUST: A Multilingual Student-Teacher Learning approach for low-resource speech recognition }
\name{Muhammad Umar Farooq, Rehan Ahmad, Thomas Hain \thanks{979-8-3503-0689-7/23/\$31.00 \copyright 2023 IEEE}}
\address{Speech and Hearing Research Group, University of Sheffield, UK}
\begin{document}
%

\maketitle

\begin{abstract}

Student-teacher learning or knowledge distillation (KD) has been previously used to address data scarcity issue for training of speech recognition (ASR) systems. However, a limitation of KD training is that the student model classes must be a proper or improper subset of the teacher model classes.
It prevents distillation from even acoustically similar languages if the character sets are not same. 
In this work, the aforementioned limitation is addressed by proposing a MUltilingual Student-Teacher (MUST) learning which exploits a posteriors mapping approach.
A pre-trained mapping model is used to map posteriors from a teacher language to the student language ASR. These mapped posteriors are used as soft labels for KD learning.
Various teacher ensemble schemes are experimented to train an ASR model for low-resource languages. A model trained with MUST learning reduces relative character error rate (CER) up to 9.5\% in comparison with a baseline monolingual ASR.

\end{abstract}
\begin{keywords}
multilingual, knowledge distillation, automatic speech recognition, low-resource languages
\end{keywords}
%
\section{Introduction}
\label{sec:intro}

State-of-the-art automatic speech recognition models nowadays require huge amounts of data for training. However, only 23 out of 7000 language are spoken by more than half of the world’s population \cite{ethnologue}. Thus a large number of languages lack enough data resources to train a modern ASR system. Multilingual and cross-lingual systems have got a lot of attention in recent years to exploit resources of other languages to overcome the data scarcity issue for training of speech technologies for low-resource languages \cite{abate20,tachbelie20,martin16,besacier14}
. Although multilingual ASR systems are considered to perform better when compared with their monolingual counterparts of low-resource languages, the performance of these systems often degrades due to mixing of unrelated languages \cite{Pratap2020,hou20, neeraj21}. This has given rise to various studies with an aim to improve a monolingual ASR using multilingual or cross-lingual resources rather than training a unified model \cite{xu22b_interspeech, klejch22_interspeech, morshed22_interspeech}.  Recently, some efforts have been made towards multilingual knowledge distillation where multilingual models are used for knowledge distillation to train a language-specific student ASR model \cite{leal21}.

Student-teacher training or knowledge distillation (KD) \cite{hinton15} has widely been used to distil the knowledge from either a single or multiple teacher models \cite{huang2023} to train a student model. This technique of transferring a teacher's knowledge to a student model either at output layer \cite{hinton15} or at intermediate stages \cite{romero15} has been used for many tasks such as model compression \cite{huang2023, kim2019} and domain generalisation \cite{wang2021, kim2021domain, fang2021}. The student model is trained with a combined objective of minimising the KL-divergence loss for prediction of the teacher's posteriors (soft labels) and a classification loss with the original training labels (hard labels).

Since KL-divergence loss is used as KD loss between a teacher's soft labels and the student model posteriors \cite{hinton15}, the output classes of student model must be a subset of the teacher model. Studies on multilingual knowledge distillation have used teacher models where student model output classes are an improper subset of the teacher model classes \cite{leal21}. Nevertheless, it still constrains a teacher model to cover all the student classes yet a lot of languages have diverse character sets and writing scripts. As it happens, a number of languages which are acoustically similar or belong to same language families are written in different scripts such as Turkish and Kazakh (Turkic), Urdu and Hindi (Indo-Aryan), and Greek and Armenian (Indo-European). It prevents various languages to distil their knowledge for training of a closer language ASR model. Though a lot of previous works have explored knowledge distillation for domains where student and teachers are from same language and have same output classes, to the best of authors knowledge no work has been done for either cross-lingual knowledge distillation or to overcome the aforementioned problem. This paper presents a step towards overcoming the obstacle for applying KD in cross-lingual settings.

To that end, a posteriors mapping technique is exploited here which has recently been proposed with an objective to analyse the cross-lingual acoustic-phonetic similarities \cite{farooq22a}. A mapping model has been trained to map posteriors from a source language ASR to those of a target language ASR given a target language speech utterance. The work has been employed for multilingual and cross-lingual model fusion for speech recognition on phonemes level \cite{farooq22b} and end-to-end ASR systems \cite{farooq23b}. In this work, we make use of a source (teacher) language ASR model followed by a source-target (teacher-student) mapping model to act as a teacher for student model training. Source and target are synonymously used as teacher and student respectively for rest of the paper. For $N$ languages, one mapping model is trained for each target language to map posteriors from other $N-1$ source languages ASR models to the posteriors of the target language ASR. Posteriors from ASR model of a source language followed by the target language mapping model are used as soft labels for knowledge distillation. Having multiple teachers from $N-1$ source languages, different existing weighting schemes along with a proposed self-adaptive weighting (SAW) are experimented for teachers ensemble to generate soft-labels. The key contribution of MUST learning is to overcome the limitation of multilingual KD and use teachers from diverse languages for multilingual knowledge distillation. ASR models trained with MUST learning for low resource languages yield a gain of up to 9.5\% in terms of relative character error rate (CER).
    

\section{Mapping models}
\label{sec:mns}

Let the monolingual acoustic models of the target language ($M_{A}$) and the $i^{th}$ source language ($M_{S_{i}}$), a mapping model $N_{S_{i}A}$ is trained to map posteriors from $M_{S_{i}}$ ($P^{S_{i}}$ of dimension $d_{S_{i}}$) to the posteriors of $M_{A}$ ($P^{S_{i}A}$ of dimension $d_{A}$). Given a set of target language observations \(X=\{x_{1},x_{2},\dotsc,x_{T}\}\), posterior distributions from the target and the $i^{th}$ source acoustic models (\(P^{Z}=\{p_{1}^{Z},p_{2}^{Z},\dotsc,p_{T}^{Z}\}\) where \(Z \in \{A,S_{i}\} \)) are attained. A sequence-to-sequence mapping model is trained to map posteriors from $i^{th}$ source acoustic model ($P^{S_{i}}$) to the target language posteriors ($P^{S_{i}A}$) using the KL-divergence loss as follows;
\begin{equation}
\label{eq:kl}
\mathcal{L}_{S_{i}A}(\theta)=\sum_{n=1}^{B} p^{A}_{n} \cdot (\log p^{A}_{n}- \log p^{S_{i}A}_{n})
\end{equation}
where $B$ is the number of frames in one batch for training of a mapping model. 

Mapping models are trained to learn the mappings between posterior distributions from a source language and the target language ASR given a target language utterance. An underlying assumption is that these mapping models are able to learn some language-related relationships between posterior distributions of a source and the target language acoustic models.

\begin{figure}[t]

    \centering
    \includegraphics[width=\linewidth]{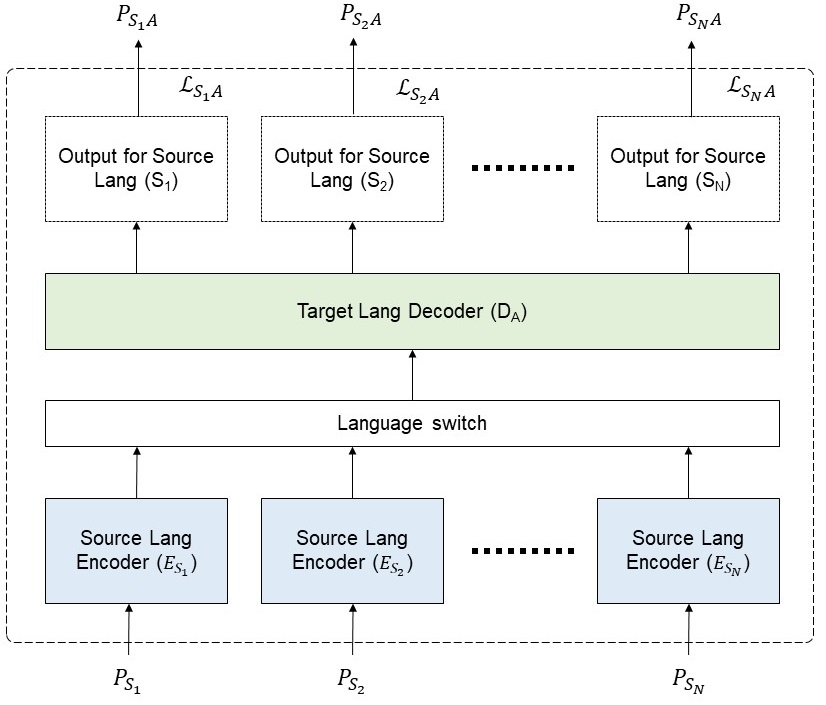}
    \caption{Architecture of the MESD mapping model \cite{farooq23b}}
    \label{fig:archi}
    
\end{figure}

Multi Encoder Single Decoder (MESD) architecture, as proposed in \cite{farooq23b} and shown in Figure \ref{fig:archi}, is used for all the mapping models.
A single MESD model is trained for each target language which consists of multiple encoders (same as number of source languages) and a single decoder with a language switch in-between.

For training of the MESD mapping model, outputs from all the source acoustic models ($P^{S_{i}}$) for a given utterance $u$, are fed to source-language dependent encoders successively. Embeddings from the final layer of the encoders are then passed to a single target-language dependent decoder. Target posteriors ($P^A$) are generated by decoding utterance $u$ through the target-language ASR. KL loss (Equation \ref{eq:kl}) is calculated between target posteriors ($P^A$) and the output of mapping model decoder ($P^{S_{i}A}$). Mapping model loss is calculated as the mean of the losses of all encoder-decoder pairs.
\begin{equation}
\label{eq:kl2}
\mathcal{L}_{A}(\theta)=\sum_{K} {w_{k} \cdot \mathcal{L}_{S_{k}A}}
\end{equation}
where $K$ is the number of the source languages ($N-1$). In the case of mean average, $w_k$ is given as  $w_{k}=\frac{1}{K}$. $\mathcal{L}_{S_{k}A}$ is given in Equation \ref{eq:kl} where each frame serves as a training example. It enables the mapping models to converge in low-resource setting as a small amount of data provides millions of training examples. Since the average loss of all encoder-decoder pairs for a mapping model causes unbalanced training across languages, rank sum dynamic weighting scheme \cite{Roszkowska2013} is applied to weight the losses for each encoder-decoder loss. In this scheme, the weights are assigned based on their normalised ranks. $w$ in Equation \ref{eq:kl2} then becomes
\begin{equation}
\label{eq:kl3}
w_{r}=\frac{2(K+1-r)}{K(K+1)}
\end{equation}
where $r$ is rank of the language when the source languages are sorted in descending order of their losses. It restricts model from biasing towards a specific language or a group of languages.

Though a mapping model contains multiple encoders, any encoder can be used with decoder during decoding and MESD does not require data stream from all the encoders for a given utterance. It implies that mappings can be obtained having input even from only one source language at a time. Training of these mapping models allows to use any source language ASR for decoding the data of a target language followed by the source-target mapping model. 

\section{Multilingual Student-Teacher (MUST) Learning}
\label{sec:must}

As described in the Section \ref{sec:intro}, output classes of the student model are required to be a proper or improper subset of the teacher model classes for knowledge distillation. It prevents a teacher language to distil its knowledge to train a student model if writing scripts or character sets are not the same. In this work, mapping models are employed to overcome this issue and distil knowledge from diverse teacher languages to train a student model of a low-resource language.

For a given target language ($L_{tgt}$), an encoder-decoder sequence-to-sequence monolingual acoustic model is trained using hybrid CTC loss as given in Equation \ref{eq:asrloss}.
\begin{equation}
\label{eq:asrloss}
\mathcal{L}_{ASR}(\theta)=\alpha \mathcal{L}_{CTC}+(1-\alpha)\mathcal{L}_{seq}
\end{equation}
where $\mathcal{L}_{CTC}$ is applied on top of the encoder after an affine projection layer. $\mathcal{L}_{seq}$ is cross-entropy loss which is applied on the decoder's output.

\begin{figure}[b]
    \centering
    \includegraphics[width=\linewidth]{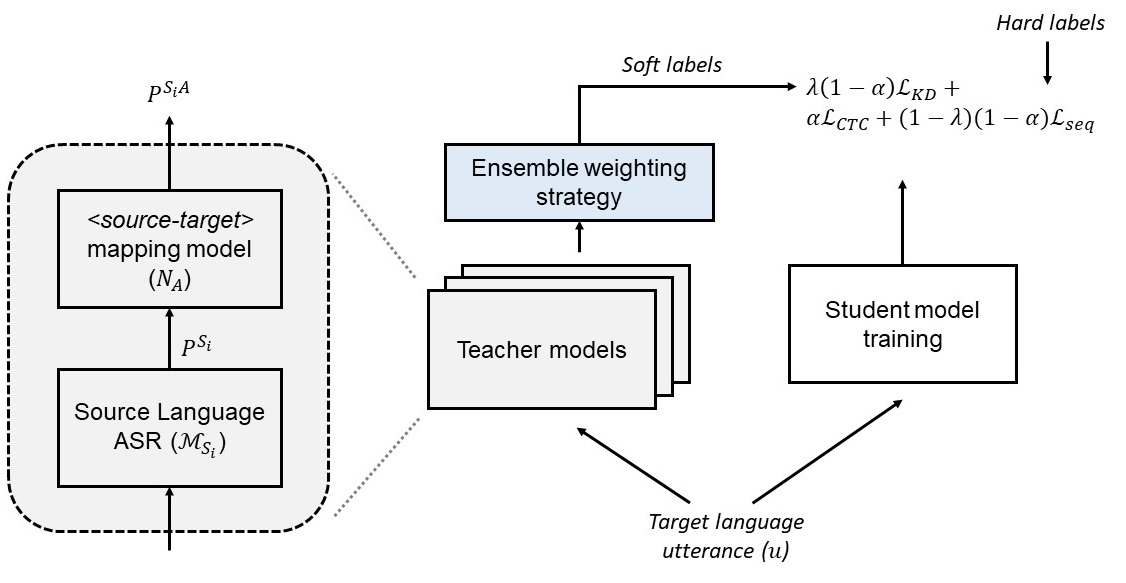}
    \caption{Architecture of Multilingual Student-Teacher (MUST) learning}
    \label{fig:mustArchi}
    \vspace{-1.5em}
\end{figure}

For MUST learning, soft-labels from a single teacher or an ensemble of multiple teacher models are used to distil knowledge for the training of a model for low-resource language. $\mathcal{L}_{seq}$ loss in Equation \ref{eq:asrloss} is modified as
\begin{equation}
\label{eq:seqloss}
\mathcal{L}_{seq}(\theta)=\lambda \mathcal{L}_{KD}+(1-\lambda)\mathcal{L'}_{seq}
\end{equation}
where $\mathcal{L'}_{seq}$ is still cross-entropy loss and $\mathcal{L}_{KD}$ is the knowledge distillation loss which is ensemble of multiple teachers and given as
\begin{equation}
\label{eq:kdlosssum}
\mathcal{L}_{KD}=\sum_{K} \mathcal{W}_{k}\mathcal{L}_{T_k}
\end{equation}
$\mathcal{L}_{T_k}$ is KL-divergence loss between posteriors from $k^{th}$ teacher model and the student model.
\begin{equation}
\label{eq:kdloss}
\mathcal{L}_{T_k}= \sum_{B} p^{T_k} \log \frac{p^s}{p^{T_k}}
\end{equation}
where $p^{T_k}$ and $p^s$ are the posterior distributions from $k^{th}$ teacher and the student model respectively. A teacher model is a source language ASR model followed by a target language mapping model $N_{A}$ as shown in Figure \ref{fig:mustArchi}. $\alpha$ and $\lambda$ in Equations \ref{eq:asrloss} and \ref{eq:seqloss} are hyper-parameters and different teacher weighting strategies are experimented for $\mathcal{W}$ in Equation \ref{eq:kdlosssum}.

Given an utterance $u$ of a target language, it is decoded through all the source language ASR systems ($M_{S_{i}}$) which generate posteriors for their output classes ($P^{S_{i}}$). Then a pre-trained target language mapping model ($N_A$) is used to map the output posteriors from source language ASR systems to the target language ASR ($P^{S_{i}A}$). Output posteriors from the mapping model are used as soft targets for student model training. So, ASR of each source language along with a target language mapping model act as a teacher model for MUST learning. For the target language student learning, experiments are conducted using an ensemble of multiple teachers (source languages) and a single teacher to generate soft labels for KD training.

\vspace{-1em}
\subsection{Self-adaptive weighting}
\vspace{-0.5em}
\label{sec:saw}
Performance of the ensemble teacher models depends on the choice of $\mathcal{W}$ in the Equation \ref{eq:kdlosssum} for each teacher loss. A straightforward approach is the teacher-averaging (TA) where all the teachers are assigned equal weights. However, all the teachers have different relationships with the student task and thus impact differently. In case of multilingual systems, all teacher languages are not equally similar and assigning the equal weights does not prove to be an optimal way.

In this work, a self-adaptive weighting (SAW) scheme is proposed. Motivated by a recent work which makes use of posterior distributions \cite{rehan23}, teacher models get relative weights based on their confidence in soft-labels. Furthermore, rather than assigning the same weights for a batch, teachers weights are calculated on-the-fly for each utterance. Given an utterance $u$ of $T$ frames, mean of $max(p_t) \forall t \in \{1,2,\cdots,T\}$ is calculated where $p_{t}$ is the posterior distribution at time $t$.

\vspace{-0.5em}
\begin{equation*}
\vspace{-0.5em}
\label{eq:meanmax}
\mathcal{\mu}_{k}= \frac{1}{T} \sum_{t} \max(p_{t}^{T_k})
\end{equation*}
Then the weight of each teacher is set to

\vspace{-0.5em}
\begin{equation}
\label{eq:teachweight}
\mathcal{W}_{k}= \frac{\tau^{\mu_k}}{\sum_{K} \tau^{\mu_k}}
\end{equation}
where $\sum_{K} \mathcal{W}=1$ and $\tau$ is a hyper-parameter for the sake of statistically significant weights distribution across the teachers. Increasing the value of $\tau$ increases the deviation of the teachers weights from the mean weight.

\vspace{-1em}
\section{Experimental setup}
\vspace{-0.7em}
\label{sec:es}

\subsection{Data set}
\label{sec:dataset}
\vspace{-0.5em}
\begin{table}[t]
\centering
\caption{Details of BABEL data sets used for the experimentation}
\label{tab:data}
 \begin{tabular}{l|cc|cc}
\hline
\hline
\multirow{2}{4em}{Lang}&\multicolumn{2}{c|}{Train}&\multicolumn{2}{c}{Eval}\\
\cline{2-3}\cline{4-5}
&\# hours&\# spks&\# hours&\# spks\\
\hline
Tamil \textit{(tam)}&59.11&372&7.8&61\\
Telugu \textit{(tel)}&32.94&243&4.97&60\\
Cebuano \textit{(ceb)}&37.44&239&6.59&60\\
Javanese \textit{(jav)}&41.15&242&7.96&60\\
\hline
\hline
\end{tabular}
\vspace{-1.5em}
\end{table}

In this work, all the experiments are conducted using the same data sets as in the previous work on mapping models \cite{farooq22b}. Four low-resource languages (Tamil (\textit{tam}), Telugu (\textit{tel}), Cebuano (\textit{ceb}) and Javanese (\textit{jav})) from the IARPA BABEL speech corpus \cite{babel} with their Full Language Packs (FLP) are used for ASR training and evaluation. Most of the BABEL data sets consist of conversational telephone speech with real-time background noises and is quite challenging because of conversation styles, limited bandwidth, environment conditions and channel. All the utterances without any speech are discarded. The details of the data sets are given in Table \ref{tab:data}.

For training of the mapping models, a subset of 30 hours is randomly selected from each BABEL language pack. This data is further split into 29 hours of train set and 1 hour of dev set.
\vspace{-0.7em}
\subsection{Student and teacher models}
\vspace{-0.5em}
\label{sec:asrs}
As described earlier, Hybrid CTC/attention architecture \cite{kim17} is used to train all speech recognition models which consists of three modules that are; a shared encoder, an attention decoder and a CTC module. The training process jointly optimises the weighted sum of CTC and attention model as given in Equation \ref{eq:asrloss} but $\mathcal{L}_{seq}$ is a cross-entropy loss for the training of teacher models which implies that $\mathcal{L}_{seq}$ in Equation \ref{eq:asrloss} is same as $\mathcal{L'}_{seq}$ of Equation \ref{eq:seqloss}.

The input to the model are 40 filterbanks and the output of the model is byte-pair encoded (BPE) tokens. All the models are trained for 100 BPE tokens for each language and SentencePiece library \cite{sentencepiece} is used for tokenisation. Both student and teacher models are of the same capacity ($\sim{170.9}$ million) throughout the experimentation.

During decoding, the final prediction is made based on a weighted sum of log probabilities from both the CTC and attention components. Given a speech input $X$, the final prediction $\hat{Y}$ is given by;
\vspace{-0.7em}
\begin{equation}
\vspace{-0.5em}
\label{eq:decode}
\hat{Y} = \argmax_{Y \in \mathcal{Y}} \{\gamma \log P_{CTC} (Y|X) + (1-\gamma)\log P_{seq}(Y|X)\}
\end{equation}
where $\gamma$ is a hyper-parameter.

For speech recognition task, results are reported in terms of percent character error rate. The SpeechBrain toolkit \cite{speechbrain} is used for training of all ASR systems.

\vspace{-0.7em}
\subsection{Mapping models}
\vspace{-0.5em}
\label{sec:mapnet}

A multi encoder single decoder model is trained for each target language. In an MESD model, there are three encoders and only one attention decoder. 
Each encoder and single decoder consists of one bidirectional RNN layer. Mapping models are also of the same capacity ($\sim2.59$ millions) for all the languages and trained on equal amounts of data.

\subsubsection{Performance metric}
\label{sec:pm}
Performance of mapping models is reported in terms of accuracy. Accuracy of a mapping model is measured as the ratio of correctly mapped frames ($CMF$) to the total number of frames ($TF$). 
Correctly mapped frames are defined as the frames where the most probable classes from mapping model and the target posteriors are the same (that is $\arg \max_k(p^{A}_{t,k})==\arg \max_k(p^{S_{i}A}_{t,k})$ where $k$ is the index of a class in the output vector $p_{t}$).



\vspace{-1em}
\subsection{MUST learning}
\vspace{-0.5em}
\label{sec:mustlearn}
For the experimentation in this work, values of $\alpha$ and $\lambda$ of Equation \ref{eq:asrloss} and \ref{eq:seqloss} are varied between the range of $[0,1]$ and the numbers are reported with the best configuration. The values of $\alpha$ and $\gamma$ are kept constant for all the experimentation while $\lambda$ may vary for different languages. For teachers' ensemble, various weighting strategies are experimented to assign the weights ($\mathcal{W}$). Conventional teacher averaging (TA) is compared with proposed self-adaptive weighting (SAW). In teacher averaging, all the teachers get the equal weights and does not change during whole training. Frame-wise max (FWM) selects posteriors from a different teacher for each frame of a given utterance. For each frame, the teacher having a maximum value of posteriors among all the teachers is selected. Recently, an elitist sampling (ES) has been proposed and prove to outperform TA and FWM weighting strategy for speech recognition domain generalisation task \cite{rehan23}. ES takes mean of maximum posterior values of all the frames for a given utterance. Then the soft labels of the teacher having the highest value are used for that given utterance. 

In previous work, posterior distributions from all the mapping models have been fused as an acoustic model which have outperformed the monolingual acoustic models \cite{farooq22b}. However, the fused weights have been fine-tuned for test set. An experiment is also done here by assigning fine-tuned weights (FTW) for the test set. These weights are manually fine-tuned and might be a sub-optimal solution. Lastly, a comparison is shown with using only one teacher model for knowledge distillation rather than ensemble of all the teachers to reduce the computational complexity. The objective is to analyse the gap in performance by reducing the teacher models. In case of single teacher (ST) distillation, only the teacher from the closest language is selected. `Closest' language is defined in terms of mapping models accuracy. For a target language, the source language with maximum mapping model accuracy is selected as the teacher model.

\vspace{-1em}
\section{Results and Discussion}
\vspace{-0.5em}
\subsection{Teacher models}
\vspace{-0.5em}
As described in Section \ref{sec:must}, a teacher model for MUST learning is a combination of a teacher language ASR and a student-teacher mapping model. Since most of the languages included in this study have different scripts and character sets, ASR of a language cannot be used for decoding the data of the other one. Pre-trained mapping models are used for each student-teacher pair in this work. Performance of the pre-trained mapping models is tabulated in Table \ref{tab:mapAccu} in terms of accuracy. For each target language, accuracy of the mapping model is shown for all the source-target mapping modules.

\begin{table}[t]

    \centering
    \caption{Accuracy of the pre-trained mapping models }
    \label{tab:mapAccu}
    \begin{tabular}{lcccc}
    \hline \hline
        \multirow{2}{4em}{Source Lang.}&\multicolumn{4}{c}{\textit{Target/Student languages}}\\
        \cline{2-5}
        &\textit{tam}&\textit{tel}&\textit{ceb}&\textit{jav}\\
        \hline
        
        \textit{tam}&-&48.88&60.53&62.24\\
        \textit{tel}&47.46&-&48.32&54.64\\
        \textit{ceb}&45.98&46.22&-&65.51\\
        \textit{jav}&46.97&47.40&65.04&-\\
    \hline \hline
    \end{tabular}
    \vspace{-1.5em}
\end{table}

\begin{table}[b]
\vspace{-1em}
    \centering
    \caption{MUST teachers performance in terms of \%CER}
    \label{tab:cross}
    \begin{tabular}{lccccc}
    \hline \hline
        \multirow{2}{4em}{MUST teachers}&\multicolumn{4}{c}{\textit{Target/Student languages}}&\\
        \cline{2-5}
        &\textit{tam}&\textit{tel}&\textit{ceb}&\textit{jav}&\textit{avg}\\
        \hline
        \textit{ES}&57.24&83.23&72.09&75.93&72.12\\
        \textit{FWM}&57.39&82.54&62.45&70.14&68.13\\
        \textit{SAW}&57.34&84.31&61.99&67.32&67.74\\
        \textit{TA}&57.38&84.31&61.98&67.26&67.73\\
        \textit{FTW}&57.34&83.36&60.03&59.45&65.04\\
    \hline \hline
    
    \end{tabular}
    
\end{table}
\vspace{-0.5em}
\subsection{MUST learning}
\vspace{-0.5em}
For multilingual student-teacher learning, various teacher ensemble strategies are explored.
Before training a student model, the ensemble strategies are applied for teacher models fusion. For a given target language, outputs from all the teacher models are fused together in a weighted sum. CER is calculated by applying greedy search on fused teacher outputs. All the discussed ensemble strategies (in Section \ref{sec:mustlearn}) including teacher averaging (TA), frame-wise max (FWM), elitist sampling (ES), self-adaptive weighting (SAW) and fine-tuned weights (FTW) are experimented and results are tabulated in Table \ref{tab:cross}. 

Analysis shows that the trend of student model performance with different ensemble strategies is same as the trend for model fusion. So, the student models here are trained using only top three best performing teachers' ensemble techniques in Table \ref{tab:cross} which are SAW, TA and FTW. \%CER of student model are shown in Table \ref{tab:must}. First row is the \%CER from a baseline monolingual ASR using an explicit RNNLM trained on limited text of train set transcriptions. Although the performance of TA and SAW is almost same for model fusion (in Table \ref{tab:cross}), student models trained with SAW ensemble reduces average CER by 1.27\% relative if compared with the models trained with TA weighting (Table \ref{tab:must}). With the weights fine-tuned for test set, average CER is reduced to 42.13\% from 42.93\% of SAW trained models which is a relative improvement of 2.4\% compared to monolingual models.

In another experiment, knowledge from only a single teacher model is distilled for student model training (ST in Table \ref{tab:must}). For each target language, the closest language is chosen as a teacher model. As described earlier, a closest source language for a target language is the one which has highest mapping model accuracy for the target language. Student models trained using the single teacher outperform all other students for all the languages by an average improvement of 4\% in performance of monolingual model. For \textit{jav} target language, a relative improvement of 9.5\% is observed.

Both \textit{ceb} and \textit{jav} yield more gains in performance than \textit{tam} and \textit{tel} because the mapping models' accuracies are higher for these two languages. It is evident that the gain for each language depends directly on the performance of corresponding mapping model. Student training with ST does not have any test set information and performs even better than FTW which has fine-tuned weights for the test set.
The results are inline with the performance of mapping models and the results reported for data augmentation using the mapping models in \cite{farooq23b}. Since some source-target mapping models does not perform very well for some of the teacher languages, the teacher knowledge introduces noise in student training and makes it hard for student to learn.
Knowledge distillation from only a single student not only improves ASR performance but also reduces the computational complexity.
\begin{table}[t]

    \centering
    \caption{Performance (\%CER) of student model trained using MUST learning}
    \label{tab:must}
    \begin{tabular}{lccccc}
    \hline \hline
        \multirow{2}{4em}{MUST teachers}&\multicolumn{4}{c}{\textit{Target/Student languages}}&\\
        \cline{2-5}
        &\textit{tam}&\textit{tel}&\textit{ceb}&\textit{jav}&\textit{avg}\\
        \hline
        
        \textit{mono}&44.28&56.18&31.26&40.90&43.16\\
        \textit{TA}&44.72&57.02&32.43&42.61&44.20\\
        \textit{SAW}&44.59&56.14&31.87&39.11&42.93\\
        \textit{FTW}&44.42&55.79&30.80&37.50&42.13\\
        \textit{ST}&\textbf{43.77}&\textbf{55.56}&\textbf{29.43}&\textbf{36.98}&\textbf{41.44}\\
    \hline \hline
    \vspace{-3em}
    \end{tabular}
    
\end{table}
\vspace{-1em}
\section{Conclusion}
\vspace{-0.5em}
This paper presents a multilingual student-teacher (MUST) approach to address a limitation of knowledge distillation systems to apply in a cross-lingual settings.
In MUST learning, a teacher model is a combination of a source language ASR followed by a source-target mapping model. Pre-trained mapping models are used to map posteriors from a source language ASR to those of the target language ASR (Table \ref{tab:mapAccu}). 
Various weighting strategies are explored for teachers ensemble (Table \ref{tab:cross}). Student models are trained for each language with top performing ensemble strategies. A student model trained with MUST learning proves to outperform baseline monolingual ASR by a relative gain of up to 9.5\%.

\vspace{-1em}
\section{Acknowledgements}
\vspace{-0.5em}
This work was partly supported by LivePerson Inc. at the Liveperson Research Centre.


\bibliographystyle{IEEEtran}
\ninept
\bibliography{refs}

\begin{thebibliography}{10}
\providecommand{\url}[1]{#1}
\csname url@samestyle\endcsname
\providecommand{\newblock}{\relax}
\providecommand{\bibinfo}[2]{#2}
\providecommand{\BIBentrySTDinterwordspacing}{\spaceskip=0pt\relax}
\providecommand{\BIBentryALTinterwordstretchfactor}{4}
\providecommand{\BIBentryALTinterwordspacing}{\spaceskip=\fontdimen2\font plus
\BIBentryALTinterwordstretchfactor\fontdimen3\font minus \fontdimen4\font\relax}
\providecommand{\BIBforeignlanguage}[2]{{%
\expandafter\ifx\csname l@#1\endcsname\relax
\typeout{** WARNING: IEEEtran.bst: No hyphenation pattern has been}%
\typeout{** loaded for the language `#1'. Using the pattern for}%
\typeout{** the default language instead.}%
\else
\language=\csname l@#1\endcsname
\fi
#2}}
\providecommand{\BIBdecl}{\relax}
\BIBdecl

\bibitem{ethnologue}
``Languages of the world,'' \url{https://www.ethnologue.com/guides/how-many-languages}, accessed: 2023-07-18.

\bibitem{abate20}
S.~T. Abate, M.~Y. Tachbelie, and T.~Schultz, ``Multilingual acoustic and language modeling for ethio-semitic languages,'' in \emph{Proc. Interspeech 2020}, 2020, pp. 1047--1051.

\bibitem{tachbelie20}
M.~Y. Tachbelie, S.~T. Abate, and T.~Schultz, ``Development of multilingual asr using globalphone for less-resourced languages: The case of ethiopian languages,'' in \emph{Proc. Interspeech 2020}, 2020, pp. 1032--1036.

\bibitem{martin16}
M.~Karafiát, M.~K. Baskar, P.~Matějka, K.~Veselý, F.~Grézl, and J.~Černocky, ``Multilingual blstm and speaker-specific vector adaptation in 2016 but babel system,'' in \emph{IEEE SLT}, 2016, pp. 637--643.

\bibitem{besacier14}
L.~Besacier, E.~Barnard, A.~Karpov, and T.~Schultz, ``Automatic speech recognition for under-resourced languages: A survey,'' \emph{Speech Communication}, vol.~56, pp. 85--100, 2014.

\bibitem{Pratap2020}
V.~Pratap, A.~Sriram, P.~Tomasello, A.~Hannun, V.~Liptchinsky, G.~Synnaeve, and R.~Collobert, ``{Massively Multilingual ASR: 50 Languages, 1 Model, 1 Billion Parameters},'' in \emph{Proc. Interspeech 2020}, 2020, pp. 4751--4755.

\bibitem{hou20}
W.~Hou, Y.~Dong, B.~Zhuang, L.~Yang, J.~Shi, and T.~Shinozaki, ``{Large-Scale End-to-End Multilingual Speech Recognition and Language Identification with Multi-Task Learning},'' in \emph{Proc. Interspeech 2020}, 2020, pp. 1037--1041.

\bibitem{neeraj21}
N.~Gaur, B.~Farris, P.~Haghani, I.~Leal, P.~J. Moreno, M.~Prasad, B.~Ramabhadran, and Y.~Zhu, ``Mixture of informed experts for multilingual speech recognition,'' in \emph{ICASSP}, 2021, pp. 6234--6238.

\bibitem{xu22b_interspeech}
Q.~Xu, A.~Baevski, and M.~Auli, ``{Simple and Effective Zero-shot Cross-lingual Phoneme Recognition},'' in \emph{Proc. Interspeech 2022}, 2022, pp. 2113--2117.

\bibitem{klejch22_interspeech}
O.~Klejch, E.~Wallington, and P.~Bell, ``{Deciphering Speech: a Zero-Resource Approach to Cross-Lingual Transfer in ASR},'' in \emph{Proc. Interspeech 2022}, 2022, pp. 2288--2292.

\bibitem{morshed22_interspeech}
M.~Morshed and M.~Hasegawa-Johnson, ``{Cross-lingual articulatory feature information transfer for speech recognition using recurrent progressive neural networks},'' in \emph{Proc. Interspeech 2022}, 2022, pp. 2298--2302.

\bibitem{leal21}
I.~Leal, N.~Gaur, P.~Haghani, B.~Farris, P.~J. Moreno, M.~Prasad, B.~Ramabhadran, and Y.~Zhu, ``{Self-Adaptive Distillation for Multilingual Speech Recognition: Leveraging Student Independence},'' in \emph{Proc. Interspeech 2021}, 2021, pp. 2556--2560.

\bibitem{hinton15}
\BIBentryALTinterwordspacing
G.~Hinton, O.~Vinyals, and J.~Dean, ``Distilling the knowledge in a neural network,'' in \emph{NIPS Deep Learning and Representation Learning Workshop}, 2015. [Online]. Available: \url{http://arxiv.org/abs/1503.02531}
\BIBentrySTDinterwordspacing

\bibitem{huang2023}
K.~P. Huang, T.-H. Feng, Y.-K. Fu, T.-Y. Hsu, P.-C. Yen, W.-C. Tseng, K.-W. Chang, and H.-Y. Lee, ``Ensemble knowledge distillation of self-supervised speech models,'' in \emph{ICASSP 2023 - 2023 IEEE International Conference on Acoustics, Speech and Signal Processing (ICASSP)}, 2023, pp. 1--5.

\bibitem{romero15}
A.~Romero, N.~Ballas, S.~E. Kahou, A.~Chassang, C.~Gatta, and Y.~Bengio, ``Fitnets: Hints for thin deep nets,'' 2015.

\bibitem{kim2019}
H.-G. Kim, H.~Na, H.~Lee, J.~Lee, T.~G. Kang, M.-J. Lee, and Y.~S. Choi, ``Knowledge distillation using output errors for self-attention end-to-end models,'' in \emph{ICASSP 2019 - 2019 IEEE International Conference on Acoustics, Speech and Signal Processing (ICASSP)}, 2019, pp. 6181--6185.

\bibitem{wang2021}
Y.~Wang, H.~Li, L.-p. Chau, and A.~C. Kot, ``Embracing the dark knowledge: Domain generalization using regularized knowledge distillation,'' in \emph{Proceedings of the 29th ACM International Conference on Multimedia}, ser. MM '21.\hskip 1em plus 0.5em minus 0.4em\relax New York, NY, USA: Association for Computing Machinery, 2021, p. 2595–2604.

\bibitem{kim2021domain}
B.~Kim, S.~Yang, J.~Kim, and S.~Chang, ``Domain generalization on efficient acoustic scene classification using residual normalization,'' 2021.

\bibitem{fang2021}
G.~Fang, Y.~Bao, J.~Song, X.~Wang, D.~Xie, C.~Shen, and M.~Song, ``Mosaicking to distill: Knowledge distillation from out-of-domain data,'' in \emph{Advances in Neural Information Processing Systems}, M.~Ranzato, A.~Beygelzimer, Y.~Dauphin, P.~Liang, and J.~W. Vaughan, Eds., vol.~34.\hskip 1em plus 0.5em minus 0.4em\relax Curran Associates, Inc., 2021, pp. 11\,920--11\,932.

\bibitem{farooq22a}
M.~U. Farooq and T.~Hain, ``{Investigating the Impact of Crosslingual Acoustic-Phonetic Similarities on Multilingual Speech Recognition},'' in \emph{Proc. Interspeech 2022}, 2022, pp. 3849--3853.

\bibitem{farooq22b}
M.~U. Farooq, D.~A.~H. Narayana, and T.~Hain, ``{Non-Linear Pairwise Language Mappings for Low-Resource Multilingual Acoustic Model Fusion},'' in \emph{Proc. Interspeech 2022}, 2022, pp. 4850--4854.

\bibitem{farooq23b}
M.~U. Farooq and T.~Hain, ``{Learning Cross-lingual Mappings for Data Augmentation to Improve Low-Resource Speech Recognition},'' in \emph{Proc. Interspeech 2023}, 2023.

\bibitem{Roszkowska2013}
E.~Roszkowska, ``Rank ordering criteria weighting methods – a comparative overview,'' \emph{Optimum. Studia Ekonomiczne}, no. 5(65), p. 14–33, 2013.

\bibitem{rehan23}
R.~Ahmad, M.~A. Jalal, M.~U. Farooq, A.~Ollerenshaw, and T.~Hain, ``Towards domain generalisation in asr with elitist sampling and ensemble knowledge distillation,'' in \emph{ICASSP 2023 - 2023 IEEE International Conference on Acoustics, Speech and Signal Processing (ICASSP)}, 2023, pp. 1--5.

\bibitem{babel}
M.~J.~F. Gales, K.~M. Knill, A.~Ragni, and S.~P. Rath, ``{Speech recognition and keyword spotting for low-resource languages: Babel project research at CUED},'' in \emph{Proc. 4th Workshop on Spoken Language Technologies for Under-Resourced Languages (SLTU 2014)}, 2014, pp. 16--23.

\bibitem{kim17}
S.~Kim, T.~Hori, and S.~Watanabe, ``Joint ctc-attention based end-to-end speech recognition using multi-task learning,'' in \emph{2017 IEEE International Conference on Acoustics, Speech and Signal Processing (ICASSP)}, 2017, pp. 4835--4839.

\bibitem{sentencepiece}
T.~Kudo, ``Subword regularization: Improving neural network translation models with multiple subword candidates,'' 2018.

\bibitem{speechbrain}
M.~Ravanelli, T.~Parcollet, P.~Plantinga, A.~Rouhe, S.~Cornell, L.~Lugosch, C.~Subakan, N.~Dawalatabad, A.~Heba, J.~Zhong, J.-C. Chou, S.-L. Yeh, S.-W. Fu, C.-F. Liao, E.~Rastorgueva, F.~Grondin, W.~Aris, H.~Na, Y.~Gao, R.~D. Mori, and Y.~Bengio, ``Speechbrain: A general-purpose speech toolkit,'' 2021.

\end{thebibliography}

\end{document}